# Agent-Based Modeling and Deep Neural Networks for Establishing Digital Twins of Secure Facilities under Sensing Restrictions

| Chathika Gunaratne[+] | Mason Stott[*], Debraj De[+], Gautam Malviya Thakur | Chris Young |
|---|---|---|
| **Computer Science and Mathematics Division** **Oak Ridge National Laboratory** | **Geospatial Science and Human Security Division** **Oak Ridge National Laboratory** | **Research Reactors Division** **Oak Ridge National Laboratory** |
| **Oak Ridge, TN** | **Oak Ridge, TN** | **Oak Ridge, TN** |
| gunaratnecs@ornl.gov | stottmc@ornl.gov, ded1@ornl.gov, thakurg@ornl.gov | youngcd@ornl.gov |

[+] Corresponding authors. [*]Intern at ORNL

## ABSTRACT

Digital twin technologies help practitioners simulate, monitor, and predict undesirable outcomes *in-silico*, while avoiding the cost and risks of conducting live simulation exercises. Virtual reality (VR) based digital twin technologies are especially useful when monitoring human Patterns of Life (POL) in secure nuclear facilities, where live simulation exercises are too dangerous and costly to ever perform. However, the high-security status of such facilities may restrict modelers from deploying human activity sensors for data collection. This problem was encountered when deploying *MetaPOL*, a digital twin system to prevent insider threat or sabotage of secure facilities, at a secure nuclear reactor facility at Oak Ridge National Laboratory (ORNL). This challenge was addressed using an agent-based model (ABM), driven by anecdotal evidence of facility personnel POL, to generate synthetic movement trajectories. These synthetic trajectories were then used to train deep neural network surrogates for next location and stay duration prediction to drive NPCs in the VR environment. In this study, we evaluate the efficacy of this technique for establishing NPC movement within *MetaPOL* and the ability to distinguish NPC movement during normal operations from that during a simulated emergency response. Our results demonstrate the success of using a multi-layer perceptron for next location prediction and mixture density network for stay duration prediction to predict the ABM generated trajectories. We also find that NPC movement in the VR environment driven by the deep neural networks under normal operations remain significantly different to that seen when simulating responses to a simulated emergency scenario.

## ABOUT THE AUTHORS

**Chathika Gunaratne, Ph.D.** is a Research Scientist in AI/ML at the Computer Science and Mathematics Division of Oak Ridge National Laboratory (ORNL). His research spans the study of complex systems through scalable artificial intelligence and modeling and simulation on high-performance computing systems and includes modeling and simulation of human patterns of life, scalable deep learning and neural architecture search for cancer detection, scalable agent-based simulation of energy efficient spiking neural networks, and modeling of misinformation and disease spread through complex social networks. Chathika is a recipient of the 2023 R&D 100 award and has co-authored over 35 peer-reviewed journal articles and conference proceedings. Chathika holds a Ph.D. in Modeling and Simulation from the University of Central Florida, and his prior appointments include a postdoctoral position at MIT-CSAIL, and modeling and simulation engineering appointments at NBC Universal Studios and SimCentric Technologies.

**Mason Stott** is a graduate student at the University of Tennessee, Knoxville and has been a research intern at Oak Ridge National Laboratory (ORNL) for over a year through the Oak Ridge Institute for Science and Education (ORISE). His research focuses on virtual reality and human computer interaction, and the applications of these technologies to the study of human mobility. Mason Stott has a B.S. in Computer Science with a minor in Machine Learning and is now pursuing a M.S. degree in Computer Science. He has experience in video game development, immersive simulations, and AI/machine learning.

**Debraj De, Ph.D.** is an R&D Staff Member/Scientist at Oak Ridge National Laboratory (ORNL). ORNL is one of the U.S. Department of Energy (DOE) science and energy laboratories, conducting basic and applied research to deliver transformative solutions to compelling problems in energy and security. At ORNL Dr. De works in the National Security Sciences Directorate and Geospatial Science and Human Security Division. Dr. De holds PhD in Computer Science, and has expertise in AI/machine learning, human mobility, multi-modal spatial temporal data, big data, data science, parallel and distributed computing, human dynamics, national security, patterns of life, pervasive computing, smart healthcare. Dr. De is a Senior member of IEEE.

**Gautam Malviya Thakur**, **Ph.D.** is a Senior Staff Scientist and the founding group leader of the Location Intelligence Group in the Geospatial Science and Human Security Division at Oak Ridge National Laboratory (ORNL). His research interests span interconnected topics related to activity-driven human mobility modeling, place-based characterization, multi-scale global land use modeling, passive sensing, and spatially explicit disinformation detection. Dr. Thakur is a Senior member of both ACM and IEEE.

**Chris D. Young** is a Design Basis Lead in the Research Reactors Division at Oak Ridge National Laboratory (ORNL).

This manuscript has been authored by UT-Battelle LLC under contract DE-AC0500OR22725 with the US Department of Energy (DOE). The publisher acknowledges the US government license to provide public access under the DOE Public Access Plan (https://energy.gov/downloads/doe-public-access-plan).

This research was sponsored by the Laboratory Directed Research and Development Program of Oak Ridge National Laboratory, managed by UT-Battelle, LLC, for the U. S. Department of Energy.

# Agent-Based Modeling and Deep Neural Networks for Establishing Digital Twins of Secure Facilities under Sensing Restrictions

| Chathika Gunaratne[+] | Mason Stott[*], Debraj De[+], Gautam Malviya Thakur | Chris Young |
|---|---|---|
| Computer Science and Mathematics Division Oak Ridge National Laboratory | Geospatial Science and Human Security Division Oak Ridge National Laboratory | Research Reactors Division Oak Ridge National Laboratory |
| Oak Ridge, TN | Oak Ridge, TN | Oak Ridge, TN |
| gunaratnecs@ornl.gov | stottmc@ornl.gov, ded1@ornl.gov, thakurg@ornl.gov | youngcd@ornl.gov |

[+] Corresponding authors. [*]Intern at ORNL

## INTRODUCTION

Virtual reality-based digital twin technologies are a valuable tool for monitoring, planning, and training for emergent scenarios in high-risk workplace environments. These technologies utilize a mix of physical Internet-of-Things (IoT) sensors, virtual reality (VR), and machine learning, to replicate the state of the physical environment *in-silico* and provide a means to designing and evaluating what-if scenarios and interventions that are too costly and dangerous to enact in live simulation settings.

However, the data required to train machine learning models that drive the simulation components of digital twins may not always be readily available, especially in the paradigm of secure facilities. It is common to encounter restrictions on sensor placement within secure facilities due to physical safety, cybersecurity, electromagnetic interference, radiation, or human monitoring concerns. In such situations, digital twin development may be restricted to information such as the building blueprints, office and device layout, and anecdotal evidence regarding day-to-day activities of personnel for an extended period until a strategy for sensor deployment, if even possible, is cleared and executed. Using anecdotal evidence presents its own set of challenges as the volume of data is not sufficient to directly train statistical or machine learning models to drive simulations of personnel behavior.

In this study, we address the challenge of learning and simulating human patterns of life (POL) within secure facilities in the absence of human activity sensor data, while relying on anecdotal evidence of day-to-day activities of facility personnel. This study was motivated by the actual sensor deployment challenges faced when deploying *MetaPOL*, a digital twin framework to detect and alert against insider threats to, or sabotage of, the *High Flux Isotope Reactor* (HFIR)[1], a secure nuclear reactor facility located at Oak Ridge National Laboratory. *MetaPOL's* capabilities enhance safety and security at *HFIR* by simulating facility personnel POL and detecting anomalous behavior that has the potential to lead to catastrophic events. The end-user component of *MetaPOL* consists of an immersive VR simulator, where non-player characters (NPCs) could mimic both the daily activities of regular facility personnel and those posing an insider threat. The player could then interact and conduct daily activities within the simulated environment, and the generated data could then be used to analyze the likelihood of certain activities indicating a threat to safety and security how the player and NPCs would react in an emergency scenario. Personnel POL were used to guide regular NPC behavior and were to be reconstructed using data from Bluetooth-enabled motion detection devices placed throughout the facility. However, due to security and electromagnetic interference concerns the deployment of motion sensor devices within *HFIR* must undergo rigorous and time-consuming evaluations prior to deployment.

To address the challenge of obtaining motion sensor data for *HFIR*, we resorted to generating synthetic data through simulations of an agent-based model (ABM) of the facility that was designed using the building blueprints and anecdotal evidence of personnel behaviors and work area usage. Relying on an ABM allowed for the intuitive specification of personnel behaviors as agent rules and to base these rules on distributions approximating user work times, break durations, and visit frequencies, while simulating the effects of distance and work area layout. The ABM provided the advantage of lower computational demands of a low-graphics and low-physics environment, unlike the game engine used for the immersive simulator, allowing for complex behavior rule execution and statistical calculations guiding agent actions, and for executing simulations in the order of $1 \times 10^3$ runs to generate a rich synthetic agent trajectory and work area transitions dataset. This dataset was then used to train deep learning surrogate models for next location prediction and stay duration prediction, which were inferred by the immersive simulator when determining NPC behaviors. We investigate how distinguishable the inferred NPC movements during normal operations are from a simulated emergency response scenario, where facility guests evacuate, and facility staff increase attention to work locations specific to facility maintenance under an increased level of irrationality in behavior selection.

---

[1] https://neutrons.ornl.gov/hfir

Our findings demonstrate how deep neural network surrogates could capture the human POL generated by the ABM. Furthermore, the trajectories generated by the surrogate model-driven VR simulator are statistically similar to those generated by the ABM. Finally, we show that when NPCs in the VR simulator are instructed to respond to a simulated emergency operation, we are able to see a quantifiable difference in the movement trajectories of the simulated facility personnel.

## BACKGROUND

Digital twin technology has been employed in many application areas (Singh et al., 2022; Liu et al., 2021; Botín-Sanabria et al., 2022) including aerospace, automotive, energy, construction, infrastructure, manufacturing, and retail to monitor and maintain human behavior in secure facilities. Challenges with digital twin development include insufficient sensor data from the physical world to train machine learning models (Mihai et al., 2022; Attaran & Celik, 2023). Agent-based modeling has been widely used to simulate human behavior in indoor emergency scenarios (Li et al., 2020; Gunaratne et al., 2022) Immersive simulations in virtual reality have been utilized in many high-risk applications (Arısoy & Küçüksille, 2020) with a focus on placing human users in virtual recreations of secure environments. Such studies have delved into the applications of this technology towards the development of safety and security for nuclear sites (Stansfield, 1996). Immersive simulations have also been utilized to assess the behavior of machine-learning agents in human dynamics studies (Rojas & Yang 2013). In our earlier work, we have developed a digital twin software framework for enabling secure facility managers to simulate insider threat and sabotage scenarios and generate corresponding POL datasets for research (Gunaratne et al., 2024).

## METHODOLOGY

*MetaPOL* is a digital twin system for ensuring safety and security at secure nuclear facilities and originally consisted of a system of indoor movement sensors, deep neural networks trained on the collected movement data, and an immersive VR environment with NPC movement driven by the deep neural networks. The end objective of *MetaPOL* is to provide an immersive VR environment for personnel to engage in simulated emergency operations from which data on personnel choices that may lead to undesirable consequences within the facility can be collected.

To address delays experienced in movement sensor deployment within *HFIR*, *MetaPOL*'s human movement data requirement was substituted with synthetic trajectories generated by an ABM. In the absence of mobility sensor data for *HFIR*, it was necessary to rely on anecdotal evidence on the day-to-day activities of personnel to generate POL for the non-player characters (NPCs) within the VR environment. Due to the high graphics and physics overhead of the game engine chosen to implement the VR environment. Instead, we developed an ABM of *HFIR*, where the agents behaved according to behavior rules gathered from analyzing the facility and anecdotal evidence of personnel behaviors, to generate synthetic human movement trajectories. These synthetic agent trajectories were then used to train deep neural network surrogate models that predicted next-destination and stay duration for NPCs within the VR environment. ABMs rely on the specification of individual-scale rules of behavior and simulate the interactions of individuals to generate emergent macro-scale outcomes and are thus well suited for the observational data at hand. Furthermore, the ABM allowed for quick and easy manipulation of agent behaviors and macro-scale analytics with very low graphics and physics overhead, making it a highly suitable methodology for synthetic trajectory generation.

**Agent-based modeling of trajectories under low data availability**

The ABM was developed in AnyLogic, a multi-method modeling and simulation toolkit highly suited for modeling movement within physical spaces (Borshchev, 2014). AnyLogic uses a combination of graphical programming assists and Java code. A top-level class defining the *HFIR* space and general user behavior was defined. The blueprints of the facility were used to draw the facility walls and boundaries within the top-level class. An illustration of the virtual space is provided in Figure 1a and Figure 1b. HFIR layouts have also been made available in prior research literature (Chichester et. al. 2018).

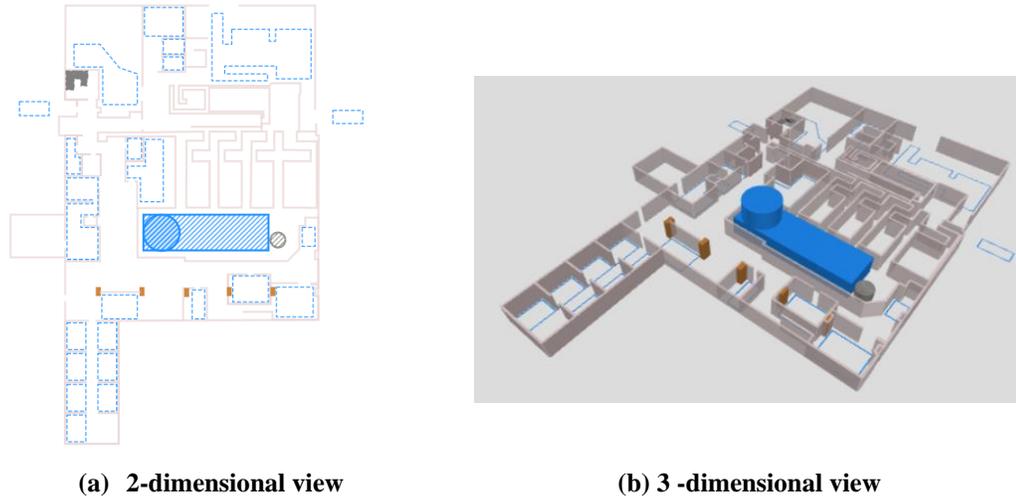

(a)  **2-dimensional view**          (b) **3 -dimensional view**

**Figure 1. 2D and 3D view of facility layout as simulated in AnyLogic.**

Anecdotal evidence was gathered through facility tours and through discussions with facility managers to understand how both the facility was used and the behaviors of the facility personnel. The following attributes were observed regarding the spaces within *HFIR*: 1) the facility consist of 2 usable entrances/exits; 2) the facility consisted of 5 categories of spaces: offices, labs, storage, maintenance, and a break room; 3) Time spent by facility personnel at offices and labs were typically 50 minutes with a minimum of 10 minutes and a maximum of 2 hours. Facility personnel typically spent 10 minutes accessing storage areas with a minimum of 5 minutes and maximum of 30 minutes. Finally, facility personnel typically spend 30 minutes in maintenance areas with a minimum of 5 minutes and maximum of 1 hour. It is important to note that these indoor spatiotemporal facility usage patterns by work groups do not necessarily follow the exact patterns at HFIR but are realistic based on anecdotal evidence. Also, some facility usage patterns of certain work groups are not used to ensure information protection.

**Table 1. Arrival schedule of all facility personnel.**

| Time Start | Time End | Percentage of Arrivals |
|------------|----------|------------------------|
| 8:00am | 8:10am | 30% |
| 8:10am | 8:20am | 30% |
| 8:20am | 8:30am | 20% |
| 8:30am | 8:40am | 10% |
| 8:40am | 8:50am | 10% |

Similarly, the follow information was observed regarding the facility personnel behaviors and was implemented as agent behaviors: 1) *HFIR* consisted of 4 classes of personnel, a facility manager (*FACILITY_MANAGER*s), facility staff or radiation workers (RAD_WORKERs), a principal investigator (*INVESTIGATOR*), and facility guest users (*FACILITY_USER*s); 2) *FACILITY_MANAGER*s and *RAD_WORKER*s typically worked 8 hour shifts, plus or minus an hour; 3) the *INVESTIGATOR* and *FACILITY_USER*s usually spent 4 hours at *HFIR*, with a minimum of 2 hours but some longer visits of up to 9 hours; 4) All users entered the facility between 8:00am and 8:50am according to the percentage arrival schedule in Table 1; 5) All facility personnel would take breaks every *inter-break duration* minutes, where *inter-break duration* would typically be 50 minutes, with a possible minimum of 20 minutes and maximum of 2 hours. Breaks would typically last 20 minutes with a minimum of 10 minutes and maximum of 1 hour; and 6) Roughly a quarter of breaks are taken outside of the facility.

Agents were designed to follow an abstract POL as illustrated in Figure 2a, with each agent deciding on actions based on its internal state machine described in Figure 2b. Specifically, agents entering the simulated facility will enter through one of the two designated entrances as specified by the *pedEnter* behavior and immediately walk to their designated work areas through the *walkToWorkArea* behavior, during which they are in *stateWorking*. Once the agent arrives at the work area, they transition to *stateWorking* by executing the *work* behavior in which they remain for a time limit sampled by the work session duration distribution based on work area type in Table 2. Agents are interrupted from this state if: 1) the time limit for *stateWorking* expires, 2) they are in time to take their next break as sampled from the inter-break duration distribution in Table 2, or 3) if their shift is over as assigned from the shift duration distribution by user class described in Table 2, at which they transition back to *movingState*. If agents are taking a break, behavior *isBreak* is executed and they may decide to exit the facility at a probability of 0.25 via the *walkOutside* behavior, or else move towards the break room via the *walkToBreakArea* behavior. The agent then remains in the *takeBreak* behavior for a time limit sampled from the break duration distribution in Table 2. After this the agent transitions into *stateMoving* once more and either: 1) returns to a designated work area through the *walkToWorkArea* behavior or 2) exits the facility if their work shift has expired. At any time, if the agent's work shift has ended, they would exit the simulation as specified by the *walkToExit* behavior and be removed from the simulation through *pedSink*.

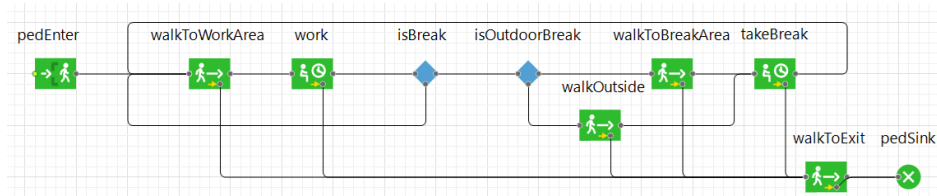

**(a) Abstract POL of agents**

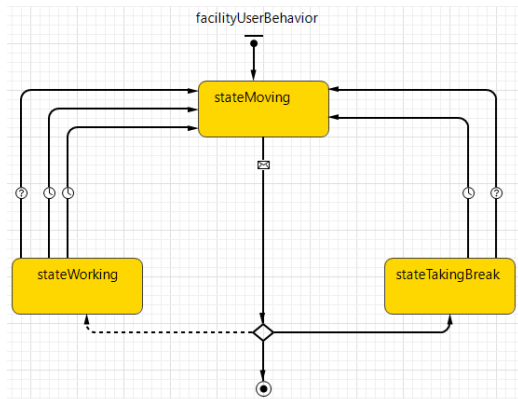

**(b) Stochastic finite state machine governing agent behavior**

**Figure 2: POL representation and agent internal state machine implemented in AnyLogic.**

1000 simulated scenarios of the ABM were run. Each run simulated a workday at *HFIR* from 07:55 AM until Midnight (12:00 AM). Each simulation was populated with a total of 16 agents, 5 facility users, 1 principal investigator, 9 facility staff, and 1 facility manager. Agents were scheduled to enter the simulation according to the entry schedule in Table 1. Both trajectories of agents within the simulated facility space and transitions between locations were logged as outputs. The trajectory outputs were recorded at 1-minute intervals and logged the agent, the agent's spatial coordinates within the facility, and current information of the agent if they were at a work or break location. Trajectories were only logged for 10 runs due to the high computational overhead of tracking the agents every minute. Transition outputs were logged whenever an agent entered the *stateMoving* state. Each transition recorded the agent, its source location, next location, and stay duration time. For each transition, location fields would be described as one of six tags: *OFFICE*, *LAB*, *STORAGE*, *MAINTENANCE*, *ENTRY*, or *END*. *ENTRY* would indicate the agent using one of the entrances for starting their shift or entering or exiting the facility during a break. *END* would indicate that the agent's shift had ended and were leaving *HFIR* for the day.

**Table 2. Parameters of triangular distributions used when specifying minutes for inter-break duration, break duration, work session duration, and shift duration under specific agent and space conditions.**

| Property | User Types | Area Types | min | max | mode |
|---|---|---|---|---|---|
| Inter-break duration | All | All | 20 | 120 | 50 |
| Break duration | All | All | 10 | 60 | 20 |
| Work session duration | All | *OFFICE*, *LAB* | 10 | 120 | 50 |
| | All | *STORAGE* | 5 | 30 | 10 |
| | All | *MAINTENANCE* | 5 | 60 | 30 |
| Shift duration | *FACILITY_MANAGER*, | All | 420 | 540 | 480 |
| | *INVESTIGATOR*, | All | 120 | 540 | 240 |

**Deep Neural Network Surrogates**

The behaviors of NPCs in the VR environment needed to be driven by data generated by the synthetic trajectories produced by the ABM. To achieve this, two deep neural network surrogates were designed and trained on the agent transitions data logged from the agent trajectories produced by the ABM simulations for 1) next location prediction and 2) stay duration at the next location.

Next destination prediction was treated as a classification task. The next destination model was implemented as a multilayer perceptron (MLP) of 3 fully connected hidden layers of sizes, 55, 110, and 55 neurons each, where each hidden layer was followed by dropout with probability 0.1 and ReLU activation functions. The network was given input features: 1) previous location type (*OFFICE*, *LAB*, *STORAGE*, *MAINTENANCE*, *ENTRY*, or *END*), 2) user class (facility manager, facility staff, facility guest, or principal investigator), and 3) the current time since the agent's first entry to the facility. Previous location type and user class were represented as one hot vectors and current time since entry was normalized. The output layer of the next destination model was a length six vector representing the six location tags. The loss function used was cross entropy loss between the predicted next location class index and actual next location class index.

Stay duration prediction was treated as a probability distribution inference task as the stay duration at a particular location was stochastic. The next destination model was implemented as a mixture density network (MDN) (Bishop, 1994; Makansi et al., 2019) with 3 fully connected hidden layers of sizes, 55, 110, and 55 neurons each, where each hidden layer was followed by dropout with probability 0.1 and hyperbolic tangent activation functions. The stay duration model used the same input feature vector as the next location model. However, during inference, the stay duration model would take the next location predicted by next destination model as output, to infer the stay duration at the next location. The output layer of the stay duration model consisted of a layer of 9 neurons, representing the mixture weights and component parameters of three Weibull distributions. The first three outputs were applied a softmax function and represented the weight vector of the samples of the three Weibull distributions, while the last six outputs represented the scale and concentration parameters of the Weibull distributions. The stay duration would be determined by sampling the three Weibull distributions described by the parameters of the last 6 outputs, multiplying the samples by their respective weights in the first 3 outputs, and summing the result. During training, loss was calculated as the negative log likelihood of the label by the 3 Weibull distribution mixture.

**Virtual reality for immersive emergency scenario generation**

A virtual reality recreation of HFIR was developed within the Unity game engine to allow human users to immersively participate in a simulation of a workday at the facility. This environment was created with the use of Unity's built-in modeling tools based on extensive reference materials and firsthand knowledge of the facility. A virtual reality player controller was created using Unity's OpenXR plugin, with a focus on support for the Meta Quest Pro. As the player moves throughout the facility, they are presented with a list of tasks they must complete in order to guide their exploration, though they can ignore these prompts should they choose to do so. All the while, the player's location data is stored by the simulator and written to a CSV file that can be downloaded off the headset for later analysis.

In addition to a human player, non-player-characters (NPCs) are also moving about the facility and following their own work schedules. These NPCs' behaviors are driven by evaluating the surrogate models with information about the individual agent that is defined in a "behavior profile," which includes information such as the agent's user class and the true location that a generic destination tag corresponds to for that individual. NPCs generate location data just as the player does.

To test black swan events in the simulator, a system was designed to throw the simulation into an *emergency mode*, in which NPCs will respond to said emergency in 2 predefined ways: 1) immediately exit the facility or 2) seek assistance from another agent. An agent's response to an emergency scenario is based on a rationality metric defined in the individual's behavior profile. Emergencies can be set to trigger at a specific time in the simulation.

## EXPERIMENTS AND RESULTS

Figure 4 visualizes the correlations of the variables describing agent transitions, specifically, source location type, user class, seconds since entry, destination location type, and destination stay duration. During inference of the deep neural networks, source location type, user class, and seconds since entry were considered as the independent variables for predicting destination location type, while user class, seconds since entry and destination location type were considered the independent variables for inferring destination stay duration. Accordingly, user class showed a high negative correlation with destination location type, while source location and seconds since entry showed mild positive correlations. Destination location type showed a high negative correlation with duration stay duration, while user class and seconds since entry showed weaker positive correlations. These observations hint that the features used to infer destination location type and duration are in fact informative. Additionally, we observed significant correlations (greater than 0.2) between the input features themselves indicating the need for a model of reasonable complexity.

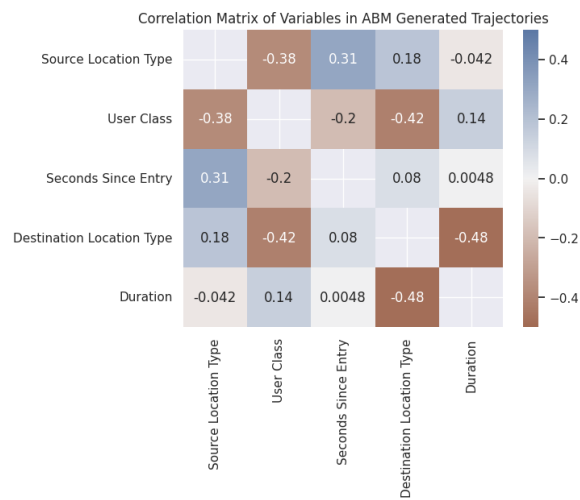

**Figure 4. Correlation matrix of variables in ABM Generated Trajectories.**

Next, we trained deep neural network surrogate models on the agent-based trajectories to predict next location and stay duration at next location. The 1000 ABM simulations produced 159032 transitions in total. The raw transition data had considerable feature imbalance which was resolved by resampling the data to produce 17000 transitions. Seconds since entry and stay duration were normalized with min-max scaling. The data was split into train, validation, and test sets of sizes 70%, 15%, and 15%, respectively. The MLP used for next location prediction was trained for 100 epochs with a batch size of 32 and learning rate 0.0001 and early stopping criteria with a patience of 3 epochs with the cross-entropy loss function. Training halted at 10 epochs due to early stopping. The metrics for the final model are provided in Table 3. The MLP was compared against a baseline model that selected the next destination type according to uniform random choice, which it outperformed at about 2.5 times higher F1 Score on the hold out test data.

**Table 3: Performance of MLP vs random baseline for next destination prediction on the test dataset.**

| Model | MAE | F1 Score | Precision | Recall | Accuracy |
|---|---|---|---|---|---|
| MLP | 1.013 | 0.522 | 0.526 | 0.520 | 0.520 |
| Baseline | 1.724 | 0.190 | 0.281 | 0.168 | 0.168 |

The MDN was trained for stay duration inference through regression on the same train, validation, and test splits above, while using current destination, user class, and seconds since entry as features and stay duration as the target. The MDN was trained for 50 epochs with a batch size of 8 and learning rate of $1 \times 10^{-5}$ and early stopping criteria with a patience of 1 epoch. Training halted at 28 epochs due to early stopping. The MDN was compared against a random baseline and a random forest regressor. The random baseline was a single Weibull distribution that was fitted to the overall distribution of stay duration of the train dataset, irrespective of user class, seconds since entry, and destination location type, and evaluated on the test dataset. The random forest regressor was trained on the same test data and used user class, seconds since entry, and destination location type as input features. The performance of the three models on the test dataset are compared in Table 4 by Wasserstein distance (also known as earth mover's distance). As shown, the MDN outperforms both the baseline and the random forest regressor.

**Table 4: Performance of MDN against the random baseline and random forest regressor for stay duration on the test dataset.**

| Model | Wasserstein Distance |
|---|---|
| MDN | $4.749 \times 10^{-3}$ |
| Baseline | $7.698 \times 10^{-3}$ |
| Random Forest | $4.396 \times 10^{-2}$ |

Next, we used the trained MLP destination and MDN stay duration models to drive the immersive Unity simulator. The Unity simulator was evaluated under two scenarios: 1) NPCs only under normal operations (*VR: NORMAL OPERATIONS*) and 2) NPCs only with emergency response (*VR: EMERGENCY RESPONSE*). Figure 5 compares the work durations of NPCs of both VR scenarios against those of the agents of the ABM. The work duration distributions of *RAD_WORKER*s and *FACILITY_MANAGER*s under *VR: NORMAL OPERATIONS* was similar to that seen in the ABM while, that under *VR: EMERGENCY RESPONSE* is significantly less. In other words, for facility staff user classes, there was a quantifiable difference in work duration when the emergency scenario is triggered in the immersive simulator. This difference was less for the *FACILITY_USER* class and absent for the *INVESTIGATOR* class. This was likely due to a combination of the lower work hours distributions specified in the ABM, and, in the case of the *INVESTIGATOR*, higher likelihood of the next destination inferred being an exit, as they had relatively less work location types to choose from.

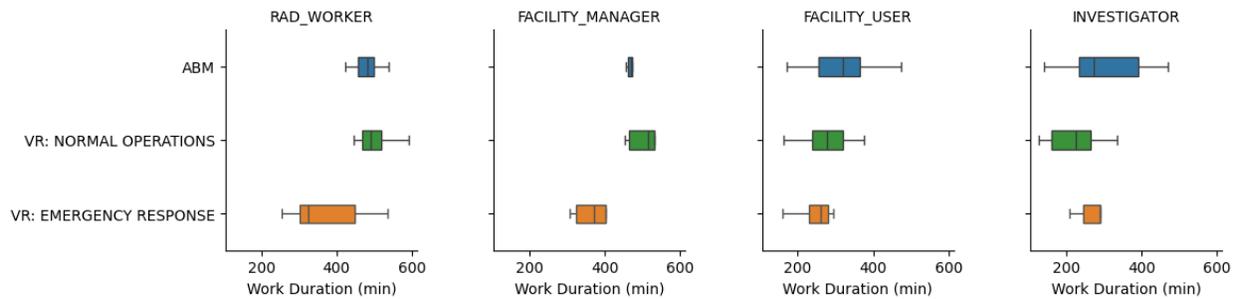

**Figure 5. Comparison of work duration of NPCs in VR scenarios and agents in ABM by user class.**

Figure 6 compares the overall trajectories of NPCs in both VR scenarios and agents of the ABM by user class over the entire simulation. Specifically, the x and y coordinates, normalized to the simulators coordinate extents are shown.

These plots ensure that the overall location visit patterns by user class generated in the AnyLogic ABM are reproduced to reasonable similarity by the Unity VR environment through the deep neural network surrogates. The general visit patterns for all classes remained the same, while discrepancies between NPC and agent walking speeds led to lower samples in the VR trajectories overall, described by the fainter lines of the second and third rows. However, the *INVESTIGATOR*s of the VR scenarios did not use the exit to the top right unlike that of the ABM, emphasizing the weakness of predicting *INVESTIGATOR* next destination described above.

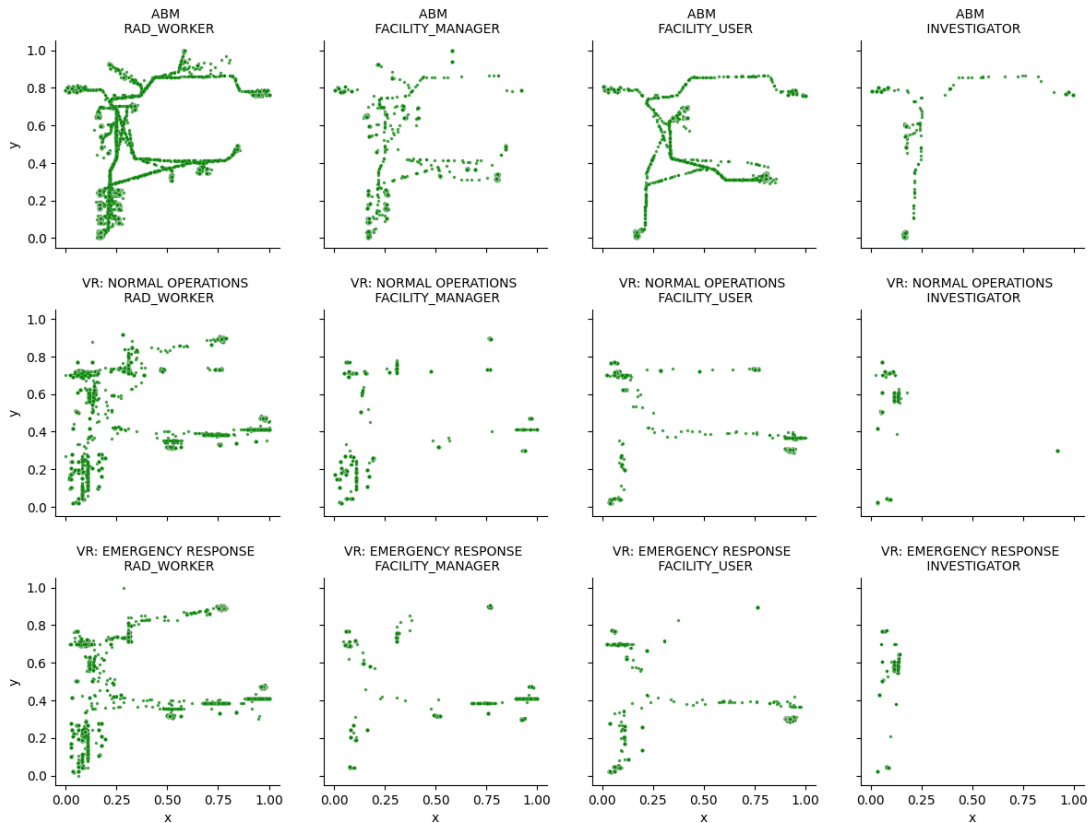

**Figure 6: Comparison of trajectories of NPCs in VR scenarios and agents in ABM by user class.**

Figure 7 compares the normalized x and y coordinates over time for all three scenarios. The horizontal dotted black line at 780 minutes indicates when the emergency scenario was triggered under *VR: EMERGENCY RESPONSE*. As expected, *FACILITY_MANAGER*s and *RAD_WORKER*s demonstrated a phase shift in their x and y coordinates following the emergency scenario trigger (orange line), while the *INVESTIGATOR*s and *FACILITY_USER*s trajectories ended as expected as they moved to exit.

Finally, to quantitatively assert the dissimilarity between *VR: EMERGENCY RESPONSE* and ABM compared to the similarity between *VR: NORMAL OPERATIONS* and ABM, we calculated the mean squared error of the mean position of agents over time by user class. Specifically, for each user class and VR scenario vs ABM comparison group, we calculated the mean squared error between normalized x and normalized y coordinates over time between the two simulators separately and then calculated the mean over both dimensions. Figure 8 confirms that the mean squared error between *VR: EMERGENCY RESPONSE* and ABM is significantly higher than that between *VR: NORMAL OPERATIONS* and ABM for the facility staff classes *FACILITY_MANAGER* and *RAD_WORKER*. Only a slight difference was seen for the *FACILITY_USER*s and the result was reversed for the *INVESTIGATOR*s, which was likely due to the emergency scenario being triggered around the time that these users ended their shift, meaning that they would have exited around that time either way.

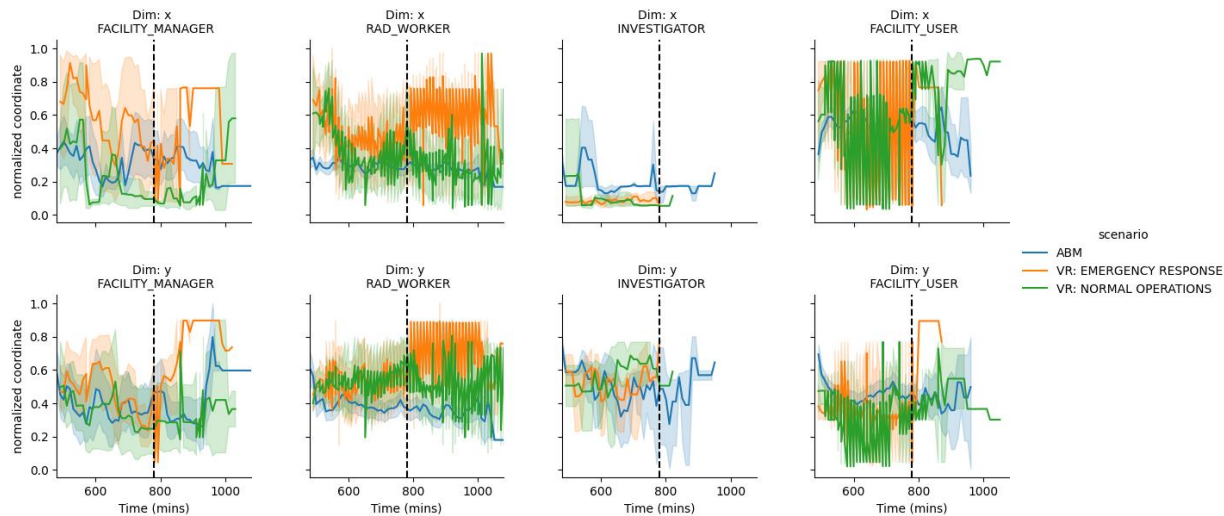

**Figure 7. Comparison of normalized coordinates over time of NPCs and agents by user class.**

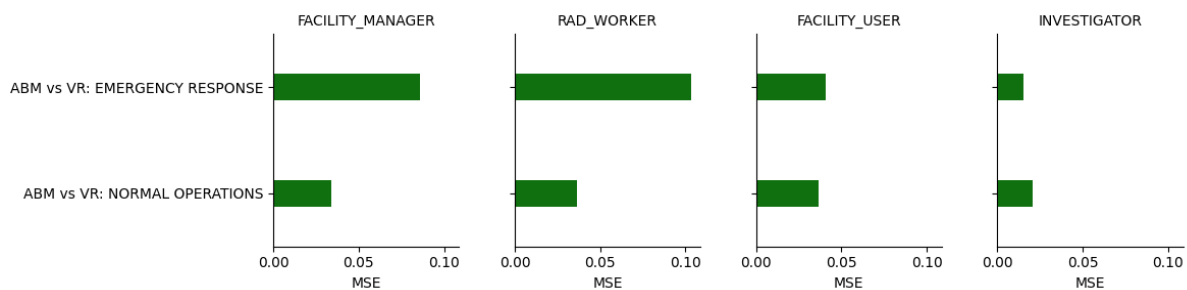

**Figure 8. Mean squared error of coordinates over time between VR scenarios and ABM by user class averaged over x and y dimensions.**

## CONCLUSION

In this paper, we demonstrate how immersive simulations for digital twin development of indoor human POL studies can be established for secure facilities under movement sensor deployment restrictions. This work was motivated by the practical challenge of establishing NPC behavior in the immersive VR simulator of the *MetaPOL* system on one of ORNL's nuclear reactor facilities, where mobility sensor deployment is restricted and has faced considerable delays. We demonstrate how anecdotal evidence of facility personnel and work location usage can be used to develop an ABM in lieu of sensor data, which can then be used as a synthetic trajectory generator. This trajectory data was then used to train deep neural network surrogates to predict both the next location and stay duration for NPCs in the VR environment. Our results demonstrate that this technique can indeed be used to drive NPCs that demonstrate significantly different POL when faced with a simulated emergency scenario. This work allows the development and evaluation of downstream components of VR based digital twins to progress despite delays in human activity sensor deployment.

## ACKNOWLEDGEMENTS

This manuscript has been authored by UT-Battelle LLC under contract DE-AC0500OR22725 with the US Department of Energy (DOE). The publisher acknowledges the US government license to provide public access under the DOE Public Access Plan (https://energy.gov/downloads/doe-public-access-plan).

This research was sponsored by the Laboratory Directed Research and Development Program of Oak Ridge National Laboratory, managed by UT-Battelle, LLC, for the U. S. Department of Energy.